\documentclass[10pt,letterpaper]{article}
\usepackage{geometry}

\usepackage{amsmath,amssymb}

\usepackage{changepage}

\usepackage{tabularx} 
\usepackage{booktabs}
\usepackage{array}

\usepackage{cite}

\usepackage{nameref,hyperref}

\usepackage{tcolorbox} 

\usepackage[nopatch=eqnum]{microtype}
\DisableLigatures[f]{encoding = *, family = * }

\usepackage[table]{xcolor}

\usepackage{array}

\newcolumntype{+}{!{\vrule width 2pt}}

\newlength\savedwidth

\newcommand\thickhline{\noalign{\global\savedwidth\arrayrulewidth\global\arrayrulewidth 2pt}%
\hline
\noalign{\global\arrayrulewidth\savedwidth}}

\providecommand{\say}[1]{``#1''}

\raggedright
\usepackage[aboveskip=1pt,labelfont=bf,labelsep=period,justification=raggedright,singlelinecheck=off]{caption}

\makeatletter
\renewcommand{\@biblabel}[1]{\quad#1.}
\makeatother

\usepackage{lastpage,fancyhdr,graphicx}
\usepackage{epstopdf}
\fancyheadoffset[L]{2.25in}
\fancyfootoffset[L]{2.25in}
\begin{document}
\vspace*{0.2in}

\begin{flushleft}
{\Large
\textbf\newline{Measurement for Opaque Systems:
Multi-source Triangulation with Interpretable Machine Learning} %
\newline
\\
Margaret J. Foster\textsuperscript{1}
\\
\bigskip
\textbf{1} National Security Data and Policy Institute, University of Virginia, Charlottesville, VA, USA
\\
\bigskip
}

* margaretfoster@virginia.edu

\end{flushleft}
\section*{Abstract}

Many high-stakes systems of scientific and policy interest are difficult, if not impossible, to reach directly: dynamics of interest are unobservable, data are indirect and fragmented across sources, and ground truth is absent or concealed. In these settings, available data often do not support conventional strategies for analysis, such as statistical inference on a single authoritative data stream or model validation against labeled outcomes.

To address this problem, we introduce a general framework for measurement in data regimes characterized by structurally missing or adversarial data. We propose combining multi-source triangulation with interpretable machine learning models. Rather than relying on accuracy against unobservable, unattainable ideal data, our framework seeks consistency across separate, partially informative models. This allows users to draw defensible conclusions about the state of the world based on cross-signal consistency or divergence from an expected state. In doing so, it
yields a defensible characterization of the system of interest.
The resulting workflow serves as a method that researchers can use to  construct, compare, and evaluate theoretical expectations about difficult-to-access targets.

\section*{Highlights}
\begin{itemize}
\item{We conceptualize measurement of adversarial systems as a triangulation problem where ground truth is unobservable or strategically concealed. We demonstrate how interpretable machine-learning models and heterogeneous data sources can produce bounded, defensible measurements in data-sparse and opaque settings.}

\item{We introduce a validation logic based on counterfactual analysis and cross-signal constraints, replacing accuracy-based evaluation with a logic of multimodel convergence and divergence.}

\item{We validate the framework in a hard empirical case and suggest portability across domains, including illicit networks, supply-chain risk, and policy analysis.}
\end{itemize}

\section*{Summary}

We propose a measurement framework for difficult-to-access contexts that uses indirect data traces, interpretable machine-learning models, and theory-guided triangulation to fill inaccessible measurement spaces. The framework provides an analytical workflow tailored to quantitative characterization in the absence of data sufficient for conventional statistical or causal inference. We demonstrate our approach and explicitly surface inferential limits through an empirical analysis of organizational growth and internal pressure dynamics in a clandestine militant organization, drawing on multiple observational signals that individually provide incomplete and biased views of the underlying process. The results show how triangulated, interpretable ML can recover substantively meaningful variation.

\section*{Introduction}

Many organizations operate under opaque and adversarial data environments in which processes are hidden, data are scarce, and incentives encourage obscuring or minimizing data outflows. Such systems produce a challenge for analysts and policymakers, who are often interested in supporting or rejecting specific theoretical expectations but lack access to reliable, fine-grained, and comprehensive data needed for traditional statistical inference. Complementary qualitative approaches often depend on rare and hard-to-replicate sources, such as insider access, defector testimony, or archival records. These sources allow ``thick'' description and qualitative inference~\cite{ponterotto2006brief, king2021designing}, but inherently limit the set of questions and cases accessible to researchers.

In this manuscript, we propose a framework for analysis in contexts where adversarial data conditions preclude statistical precision but analysts want systematic characterization. We show how computational methods on publicly available data can generate insight in a ``meso-level'' inferential space between macro-level summaries and micro-level precision (\textit{e.g.}, point estimates, confidence intervals, and causal identification). We propose a triangulation approach that uses theoretical expectations to structure expected and counterfactual performance patterns across models estimated on multiple data sources associated with the target of interest. Data-model pairings illuminate facets of our system of interest, while theory specifies patterns of performance across models. Our empirical case serves as a validation environment for the measurement strategy.

\begin{figure}[!h]
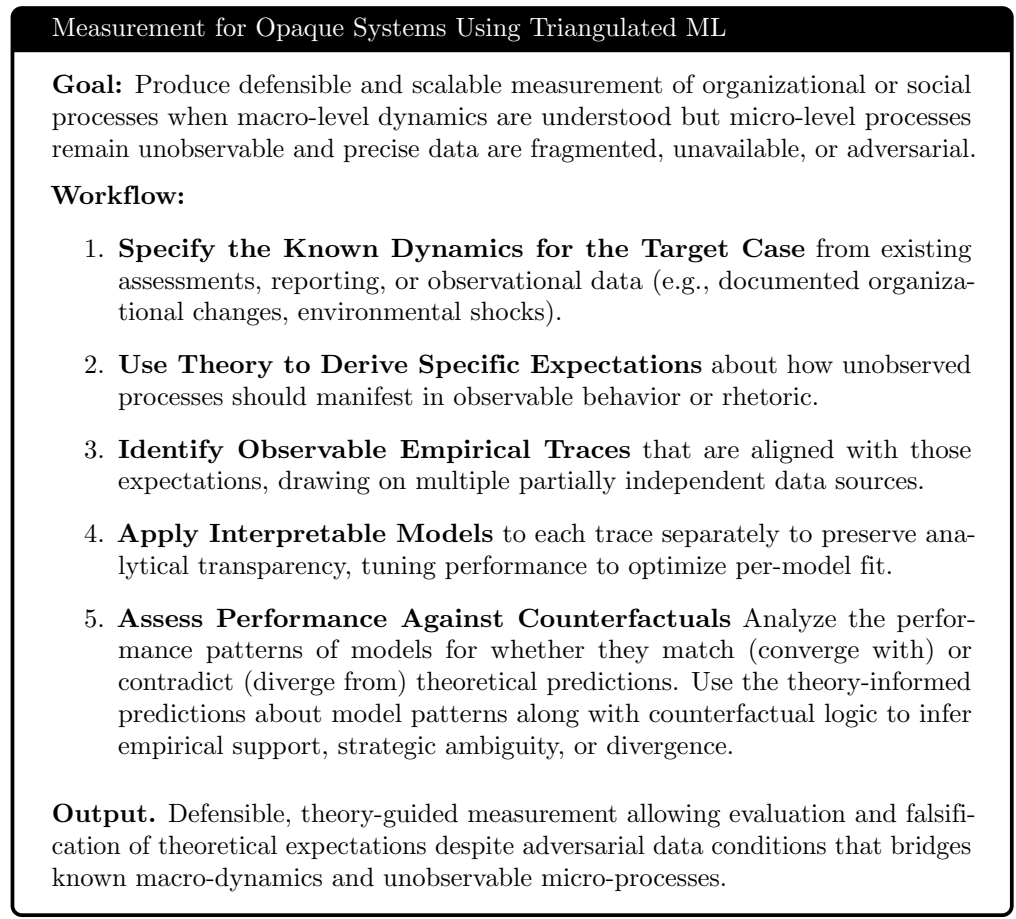

\caption{Framework Summary}
\label{fig:box}
\begin{tcolorbox}[colback=white,colframe=black,title= Measurement for Opaque Systems Using Triangulated ML]
\textbf{Goal:} Produce defensible and scalable measurement of organizational or social processes when macro-level dynamics are understood but micro-level processes remain unobservable and precise data are fragmented, unavailable, or adversarial.

\medskip
\textbf{Workflow:}
\begin{enumerate}
\item \textbf{Specify the Known Dynamics for the Target Case} from existing assessments, reporting, or observational data (e.g., documented organizational changes, environmental shocks). 

\item \textbf{Use Theory to Derive Specific Expectations} about how unobserved processes should manifest in observable behavior or rhetoric.
\item \textbf{Identify Observable Empirical Traces} that are aligned with those expectations, drawing on multiple partially independent data sources.
\item \textbf{Apply Interpretable Models} to each trace separately to preserve analytical transparency, tuning performance to optimize per-model fit.
\item \textbf{Assess Performance Against Counterfactuals} Analyze the performance patterns of models for whether they match (converge with) or contradict (diverge from)  theoretical predictions. Use the theory-informed predictions about model patterns along with counterfactual logic to infer empirical support, strategic ambiguity, or divergence.
\end{enumerate}

\medskip
\textbf{Output.} Defensible, theory-guided measurement allowing evaluation and falsification of theoretical expectations despite adversarial data conditions that bridges known macro-dynamics and unobservable micro-processes.
\end{tcolorbox}
\end{figure}

The framework proposes a measurement strategy for contexts in which data are too sparse or adversarial for traditional quantitative approaches. By using theory to derive performance expectations across a suite of models that use multiple partially independent data sources, analysts can evaluate whether observed model performance patterns match those predicted by the theoretical scaffolding or whether alternative scenarios best explain the results.  We generalize this insight into a reusable design that can be ported to other adversarial systems and enable quantitative measurement in settings where standard statistical or causal designs are structurally inappropriate. 


Through this framework, we build on the literature on interpretable machine learning to produce rigorous measurement in adversarial information environments. We draw on the literature on ML interpretability and explanatory tools~\cite{doshi2017towards,kaur2020interpreting,  rudin2019stop} to propose combining substantive domain knowledge and theory-structured counterfactuals such that model performance itself becomes diagnostic evidence. Models that successfully differentiate theoretically distinct entities validate behavioral differences. Classification failures can indicate converging activity profiles. This diagnostic interpretation of model outputs distinguishes the approach from standard predictive machine learning, where classification errors are interpreted as model inadequacy. Instead, we use errors as evidence of mismatch between the observed and hypothesized underlying system, extending the logic of work that treats errors as informative signals and distributional mismatch rather than noise~\cite{ben2010theory, suresh2019framework}. Our emphasis on tailoring evaluation to contextual features has a parallel in interpretability, which likewise context-dependent on audience and goal~\cite{doshi2017towards, lipton2018mythos, molnar2020interpretable}.
The framework provides analysts with systematic methods of deriving meaningful signals from fragmentary public data when ground-truth validation is structurally impossible. Finally, the models used in this framework are accessible to practitioners without specialized computational or mathematical infrastructure.

The primary contribution is a measurement framework, presented in Figure~\ref{fig:box}, for analysts working in adversarial information environments or studying hard-to-access organizations. Throughout, we use ``measurement'' and ``characterization'' to describe systematic, theory-guided pattern identification that produces defensible conclusions about latent processes despite pervasive uncertainty and fragmentary data. This approach occupies a distinct methodological space: it has fewer data requirements than causal identification or statistical estimation but provides more than exploratory data analysis, offering rigorous characterization when traditional inferential designs are structurally infeasible.

\begin{figure}[!h]
\centering  
\caption{{\bf Example Workflow for measurement via theory-guided triangulation.}
General framework for meso-level measurement of unobservable dynamics, applied to opaque organizational processes. Theory and prior knowledge (steps 1 and 2) constrain interpretation and generate expectations, while interpretable machine learning models are applied to multiple indirect empirical traces (steps 3 and 4). For this domain, empirical trades are derived from behavioral records, self-presentation, and external perceptions. Patterns of model performance are compared against the theory's expectations and counterfactuals (step 5), yielding theory-consistent measurement without causal identification.}
\label{fig:chart}
\includegraphics[width=.95\columnwidth]{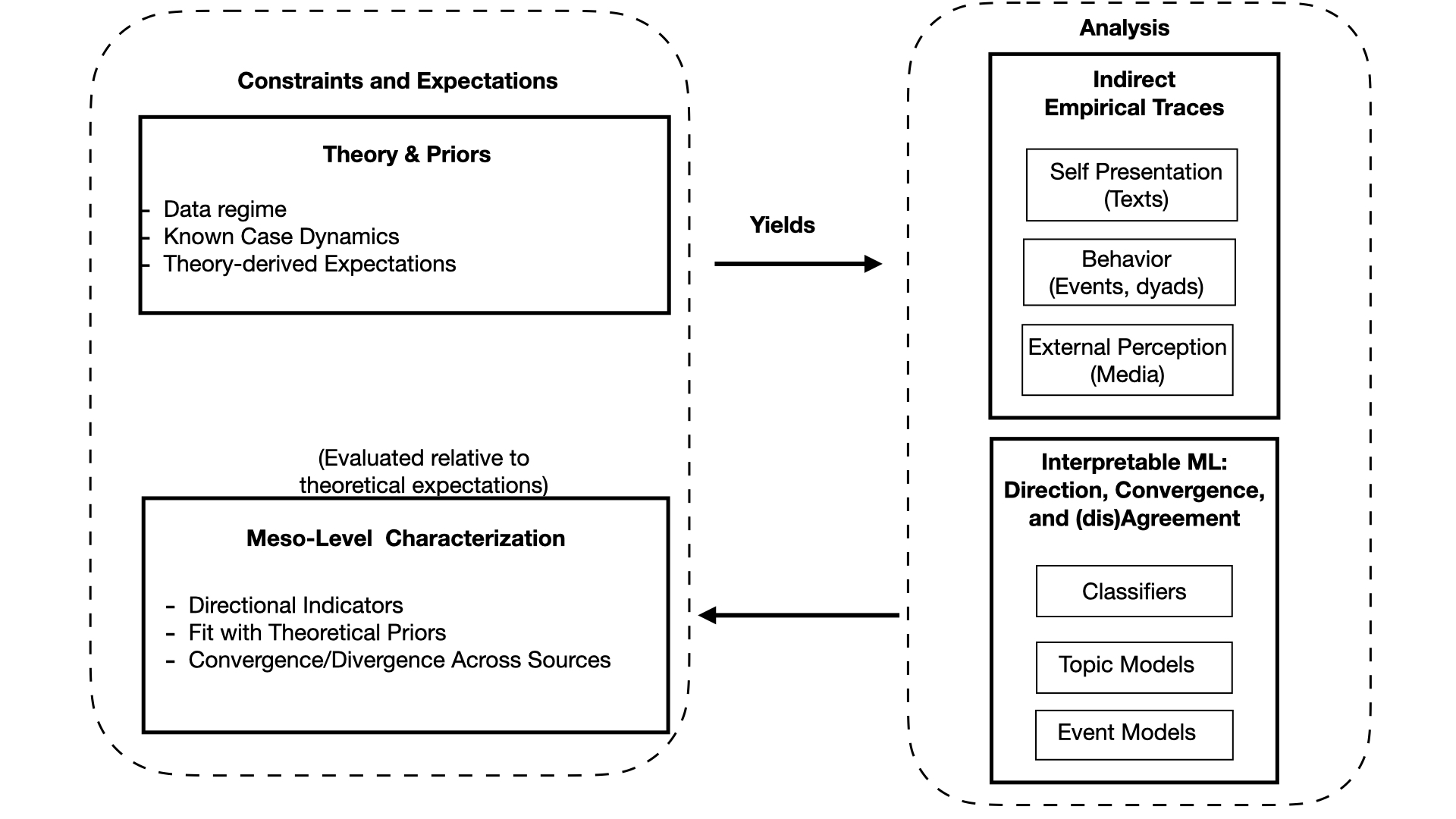}\\
\end{figure}

We demonstrate our proposed framework using al-Qaeda in the Arabian Peninsula (AQAP) because it represents a hard context for measurement. It is a high-stakes case with challenging measurement conditions. AQAP was geopolitically consequential but secretive, dangerous to access directly, and operating in environments that produce extremely sparse data. At the same time, in 2011 the group represented one of the most serious security threats to the United States~\cite{ctr2010terrorism}. The case thus embodies the data requirements and scope conditions of the measurement framework: third-party analysis records high-level dynamics, empirical traces are available, but direct access to internal processes and tensions is fundamentally unattainable.
The available evidence consists primarily of occasional qualitative reporting, indirect event data, and organizational propaganda—the kind of fragmentary, heterogeneous public data that characterizes many difficult-to-access organizations. The case shows both the promise and limitations of our strategy: we can use behavioral traces to develop predictions about performance patterns for a set of models, but require a theoretical scaffold against which to develop and evaluate expectations. Readers interested primarily in the methodological contribution rather than the substantive treatment of militant organizations can treat our Demonstration section as a worked example.
 
This work makes three contributions. First, it demonstrates how interpretable machine learning can be deployed to measure organizational phenomena that are difficult to observe directly. Second, it shows how triangulation across complementary data sources provides indirect validation absent comprehensive ground truth. Third, it presents a generalizable design applicable to other difficult-to-access organizations. We provide our workflow, all code, and discuss extensions to other domains.

\underline{Scope Conditions and Alternative Approaches}

The computational measurement framework described in this manuscript is designed for contexts with documented macro-dynamics and multiple channels of data, making it unsuitable for:

\textbf{Pure discovery}: When analysts have no baseline knowledge of organizational dynamics,  structured qualitative analysis or exploratory computational methods, such as unsupervised topic modeling and network community detection, are more appropriate.

\textbf{Organizations with minimal public traces}: Groups that successfully avoid generating observable data---such as clandestine organizations with minimal communications or groups with little externally-facing activity ---lack the multiple data channels we require.

\textbf{Environments with a single source of data}: When only one type of data exists, such as event data with no textual records, triangulation is impossible and alternative validation strategies are necessary.

\textbf{Causal inference}: Our framework is positioned between exploratory analysis and causal inference. It offers systematic measurement beyond description and can thus extend analysts' ability to measure trends and draw conclusions. However, it cannot substitute for causal identification strategies, nor can it provide estimated treatment effects.

\section*{Demonstration: Measuring Hidden Organizational Dynamics}

 In the following section, we showcase a workflow arising from our measurement framework applied to the challenge of how to evaluate the organizational dynamics of a secretive but globally consequential organization. As described earlier, AQAP represents a difficult context for measurement: the group is secretive and operates in an environment where data is difficult for most analysts to access. At the same time, as a geopolitically consequential armed organization, understanding the trajectory of AQAP is important to both regional and global actors. 

 Our triangulated design enables tracing theoretical processes across complementary types of evidence. The qualitative materials clarify ideological and organizational constraints; event data reveal behavioral shifts in target selection; and text data provide a measure of internal and external framing. Together, these sources allow a partial reconstruction of adaptation processes, an empirical indication of whether rapid growth was followed by a shift in organizational priorities, and whether any shift was likely to be sought by AQAP's top leadership.  Alternative explanations and potential sources of bias in the case are addressed in \nameref{S1_bias}.
 
\subsection*{Step 1: Specify Known Dynamics}

\begin{figure}[!h]
\centering  
\caption{{\bf Macro-Level Growth Trajectory of al-Qaeda in the Arabian Peninsula.}
A: Estimated group size for AQAP, drawn from the REVMOD dataset. B: Descriptive evolution of AQAP’s conflict dyads in Yemen, drawn from the UCDP dataset.}
\label{fig:growth}
\includegraphics[width=.85\columnwidth]{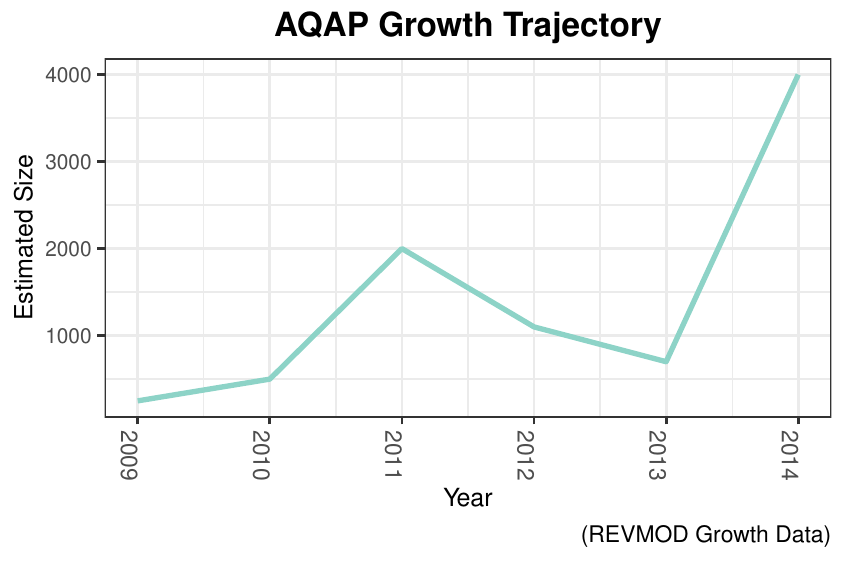}\\
\includegraphics[width=0.85\textwidth]{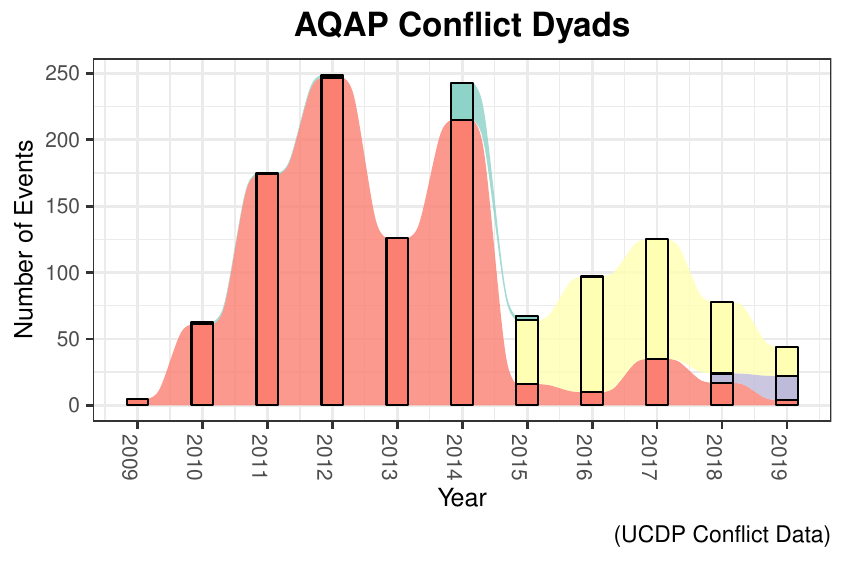}
\end{figure}

 As a starting point for measurement, we require documented macro-level dynamics. AQAP simultaneously fits multiple framework requirements: it is a closed organization with difficult-to-observe internal processes, multiple accessible external data sources exist, and growth trap theory generates specific predictions about how expansion should manifest in observable patterns. This combination makes it an effective stress test for the measurement framework. Several potential alternative cases are discussed in \nameref{S2_alternatives}. Moreover, we have documented micro-dynamics for AQAP: membership estimates from U.S. State Department assessments and the REVMOD dataset document explosive growth from approximately 200-300 members (2009) to several thousand (2011-2016), shown in Figure~\ref{fig:growth}. This known expansion provides our high-level context and anchors the timing of organizational growth against which we can evaluate our subsequent behavioral and rhetorical models. 

The AQAP case features sparse, but revealing, macro-level evidence that we can use to adjudicate between alternative explanations. Notably, captured correspondence between al-Qaeda’s central leadership and AQAP explicitly warned against local co-optation, suggesting that any shift toward parochial concerns was not centrally directed~\cite{lettertoAB}. Moreover, the group’s seizure and brief administration of territory in the Abyan Governorate in 2011, followed by territorial contraction, creates identifiable moments in which membership expansion and contraction can be linked to observable strategic shifts. In addition, AQAP’s maintenance of a semi-autonomous local affiliate, Ansar al-Shariah, provides a useful benchmark for comparison: the affiliate was explicitly designed to appeal to Yemeni local audiences\cite{kendall2018contemporary}, allowing a direct test of whether AQAP’s messaging and operations converged toward a local focus over time.

\subsection*{Step 2: Use Data to Derive Specific Expectations}

Our measurement framework requires theoretical expectations to guide model interpretation and enable triangulation. Theory provides the structures that allow our framework to draw systematic conclusions from the interplay of multiple models. 

For our AQAP motivating case, we use organizational theory to generate specific, falsifiable predictions about how AQAP's rapid expansion via local recruitment should manifest in observable behavioral and rhetorical patterns~\cite{swift2012arc, hrw2013drone, ICG2017Yemen, johnsen2012upper,koehler2011false,kendall2018contemporary}. The core prediction of our theory is that when organizations expand faster than they can socialize new members, internal coherence weakens. As a consequence, leaders face a trade-off between slowing growth to preserve ideological control or accepting preference divergence to maintain momentum. If they choose growth, the organization should shift toward the preferences of new recruits rather than founding objectives—a ``growth trap'' where expansion undermines strategic coherence.

This ``growth trap''  mechanism predicts accommodation through three channels: (1) recruitment-socialization imbalances that allow co-optation of the organizational mission, (2) exit pressures that incentivize accommodating the new preferences, and (3) delegation that transfers operational discretion to mid-level leaders who are more responsive to grassroots priorities. 

These mechanisms suggest that in the case of AQAP, rapid expansion should coincide with observable shifts in targeting behavior and rhetorical emphasis toward local Yemeni interests rather than legacy al-Qaeda transnational priorities. AQAP provides several advantages for demonstrating our framework. First, AQAP maintained a semi-autonomous local affiliate (Ansar al-Shariah) explicitly designed to appeal to Yemeni audiences, providing a natural comparison for testing behavioral convergence with locally-focused group. Second, captured correspondence between AQAP leadership and al-Qaeda's Pakistan-based central leadership (e.g. Usama Bin Laden, Ayman al-Zawahiri, and aides) explicitly warned against local co-optation. The pointed strategic instruction allows us to infer that discovered shifts represent bottom-up accommodation rather than top-down strategic adaptation. Third, AQAP's territorial gains and losses (particularly in Abyan Governorate, 2011) create identifiable moments that couple membership growth and contraction to observable behavioral shifts. Finally, the theory's predictions map cleanly onto available data sources: behavioral event data for target selection, topic models for rhetorical emphasis, and external media coverage for perceived positioning.

The ``growth trap'' theory produces testable expectations: if the mechanisms operate as expected, AQAP's twentyfold expansion (2009-2016) should produce measurable behavioral convergence with locally-oriented affiliate organizations, increased emphasis on domestic Yemeni conflicts relative to transnational jihadi themes, and temporal alignment between recruitment surges and behavioral changes. Our theory predicts which types of models we can expect to capture drift should perform well on our data (those in which AQAP controls messaging) and which should be unable to find consistent underlying patterns (empirical separation from other Sunni local conflict actors).

Our counterfactual baseline is that AQAP maintained separation from the local community. In which case, we should see AQAP continuing to present a transnational focus in the propaganda material that they control. We should also expect that our classification models using traces of revealed behavior should be able to differentiate their activity patterns from those of primarily local actors. 

We structure our expected model performance patterns from our theoretical predictions as well as under alternative scenarios. This allows us to evaluate our theory against counterfactual predictions via the ensemble of performance patterns across models. 

Given limited data on internal organizational dynamics, we use the growth trap theory outlined above as a driver of falsifiable predictions. By focusing on the observable implications of the growth-trap mechanism, we can derive specific expectations of classification models, emphasizing what the relationship between model results indicates about the system. The growth trap mechanism described above predicts that decentralization and an influx of local recruits should be followed by progressive accommodations to the operational and strategic preferences of those recruits. In the context of the Yemeni civil war, this shift is likely to manifest via an increasing focus on rival non-state and sectarian actors rather than the Yemeni or Saudi states. Similarly, its public communications should place greater emphasis on local grievances and identities rather than on al-Qaeda’s motivating transnational revolutionary agenda. Stability in these measures would instead indicate sustained leadership control and resistance to accommodating pre-recruitment preferences of the new base. We deliberately constrain model complexity to preserve interpretability and cross-source comparison. This is consistent with previous findings that transparent models often perform competitively and enable diagnostic evaluation in high-stakes settings~\cite{caruana2015intelligible, morucci2024model, rudin2019stop}.

\subsection*{Step 3: Data Sources and Traces}

The framework we present requires multiple partially independent data sources capturing different behavioral traces. As no single data source provides complete visibility, we leverage multiple datasets and data approaches. In the case of a closed organization, such as AQAP, where membership is secret, internal deliberations and pressures are unobservable, and communications may be deceptive or strategic, look for data that records intention and revealed behavioral tendencies. We combine: (1) external perception data showing how behaviors are interpreted by observers, (2) self-presentation data revealing organizational framing, and (3) behavioral event data documenting actions. 

Each stream provides incomplete insight into a facet of the behavioral record that can be triangulated to create a picture of revealed preferences and priorities. For AQAP, we identify five streams meeting these criteria, summarized below. Each of these sources provides a separate, although indirect, view into the relationship between recruitment dynamics and organizational behavior. Observational data captures the perceptions of close local observers, while official statements suggest internal priorities. 

\begin{enumerate}
\item{Qualitative Validation: Captured al-Qaeda documents, together with secondary reporting and memoirs by conflict participants, surface specific points of ideological divergence between AQAP’s founding transnational agenda and the preferences of later recruits. These materials also clarify how organizational constraints—such as the loss of training infrastructure and the delegation of authority—shape leadership control~\cite{kendall2018contemporary}}
\item{Macro-level documentation of general dynamics: Estimated fighter counts (2006-2014) establish the scale and timing of AQAP's expansion, establishing the growth pattern that triggers the accommodation mechanism. In addition to snapshot estimates published by the United States Department of State, we use the REVMOD dataset of militant group growth and organization~\cite{acosta2019reconceptualizing}.}
\item{Behavioral Traces: Geocoded conflict events document AQAP's territorial control and operational focus. These data map AQAP’s operational portfolio—measured as the proportion of violent events involving different conflict dyads (state, sectarian, tribal, and rival jihadist actors). We use the Uppsala Conflict Data Program (UCDP) Armed Conflict Dataset and extract a dataset of 1272 conflict dyads in which AQAP was a participant from 2009 - 2019\cite{sundberg2013introducing}. UCDP data on AQAP's activities are categorized by the type of opponent: the Yemeni state; the Shia-aligned Ansarullah movement (the Houthi insurgency; pro-government tribal or political actors associated with the Hadi government; and the Islamic State. The proportion of events involving each dyad provides an indicator of whether AQAP’s activities shifted from contesting the state toward engaging rival non-state and sectarian actors.}
\item{Traces of External Perception: To assess how local observers interpreted AQAP’s behavior, we use a supplementary corpus of 576 Yemeni news articles covering AQAP, its local spin-off Ansar al-Shariah, and the Houthi insurgency. The articles were sourced from the ICEWS dataset~\cite{boschee2015icews}. An ML classification of these reports provides a comparison between AQAP’s self-representation and external depictions of its priorities, allowing for evaluation across internal and external narratives.}
\item{Traces of Self-Presentation: we assess changes in AQAP's self-presentation to both local and transnational audiences via 809 professionally translated public AQAP statements issued between June 2004 and September 2016. The corpus includes material from the Yemen-based AQAP and its short-lived Saudi precursor, which AQAP’s leadership explicitly portrayed as part of a continuous organizational lineage. Through topic modeling and supervised text classification, we identify changes in rhetorical emphasis over time, distinguishing between transnational jihadist themes and locally oriented grievances or sectarian appeals.}
\end{enumerate}

\subsection*{Step 4: Interpretable ML Application}

 We turn to the fourth step of our framework: measuring patterns in data traces using statistical and machine-learning models.  First, we explicitly map how our theoretical expectations map into expected ML model behavior. Then, we implement our chosen models, maximizing per-model performance. Finally, we evaluate their performance against our expectations

The growth trap theory that motivates our application produces specific expectations about what model performance patterns should look like under its predictions, which we call the ``Accommodation'' outcome, as well as a counterfactual without drift, the ``separation'' outcome. Across all of our models, we focus on adjudicating between observable behavioral traces of global versus local focus and use qualitative evidence to determine top-down or bottom-up intent.

\textbf{Accommodation (H1):} If AQAP accommodated local recruit preferences, we predict:\\
- \textit{Random Forest}: Our highest-performing classifier should systematically fail to differentiate AQAP from Ansar al-Shariah (behaviorally convergent) while successfully differentiating both from Houthis (unrelated actor).\\
- \textit{Topic Model}: Over time, as AQAP's internal focus shifts, their public messaging should show increasing attention to local themes alongside decreasing attention to transnational jihadi themes.\\
- \textit{Event Data}: Conflict dyad proportions should shift toward local/sectarian actors after expansion periods as they accommodate the preferences of new, locally-focused, members.\\

\textbf{Separation (H2):} If AQAP maintained ideological control despite expansion, we predict:\\
- \textit{Random Forest}: Successful three-way classification distinguishing all groups.\\
- \textit{Topic Model}: Stable transnational topic prevalence regardless of membership growth.\\
- \textit{Event Data}: Consistent targeting of state actors rather than local rivals.\\

We now evaluate which performance pattern we observe.

\subsubsection*{Random Forest Classification}

 We choose models appropriate to each form of trace data, fit and refine each model separately, and then take each model as a set of evidence for evaluation.  As the data traces we have access to are drawn from written records, we use random forest and structural topic models for their performance on text-as-data. 

 We then assess model performance relative to the predictions of our theory and counterfactuals:  if the conclusions of our separate models support our expectations, we have more confidence not only in the underlying finding of organizational evolution but also in the proposed mechanism of growth-driven transformation. If, instead, they do not support our theoretical expectations, we evaluate for support for a theory-directed counterfactual. Finally, incoherent or consistently poor performance suggests that the data regime remains too thin to support inference.
 
The data and corresponding ML models are summarized in Table~\ref{tab:data_sources}.

\begin{table}[htbp]
\centering
\small
\setlength{\tabcolsep}{6pt}
\renewcommand{\arraystretch}{1.2}

\begin{tabularx}{\textwidth}{
>{\raggedright\arraybackslash}X        
>{\raggedright\arraybackslash}X        
>{\raggedright\arraybackslash}p{0.22\textwidth} 
>{\raggedright\arraybackslash}p{0.21\textwidth}        
>{\raggedright\arraybackslash}p{0.12\textwidth} 
}
\toprule
\textbf{Theory Prediction} &
\textbf{Data Source} &
\textbf{H1 (Accommodation)} &
\textbf{H2 (Separation)} &
\textbf{Model} \\
\midrule
Behavioral convergence with local affiliate &
News articles (ICEWS, $N=576$) &
Fails on AQAP/AAS separation; succeeds on Houthi differentiation &
Succeeds on all three-way classification &
Random Forest \\
Rhetorical localization over time &
AQAP statements (SITE, $N=809$) &
Local topics increase, transnational decrease post-2011 &
Stable topic proportions &
Structural Topic Model \\

Targeting shift to local actors &
Event dyads (UCDP, $N=1272$) &
Increasing local/sectarian dyad proportion &
Stable state-actor targeting &
Descriptive \\

\bottomrule
\end{tabularx}
\caption{\textbf{Data Sources, Models, and Analytic Purpose.} This table summarizes the datasets, their sources, time periods, and their corresponding analyses and purposes. For each prediction, we specify what model performance should look like under accommodation (H1) vs. maintained separation (H2) scenarios. Observed patterns support H1.}
\label{tab:data_sources}
\end{table}

The underlying theory expects that AQAP's activities should reflect local priorities after the 2011 and 2013 periods of rapid growth.
We operationalize this by leveraging third-party media coverage and the parallel operation of the Ansar al-Shariah local spin-off. If local priorities dominated, reporting on AQAP and Ansar al-Shariah should converge, despite AQAP's evident interest in separating perception of the entities~\cite{kendall2018contemporary}. At the same time, it should be possible to differentiate Sunni militias from the Houthis, an unrelated rival rebel movement operating in Yemen at the same time. 
 
 To assess whether AQAP's behavioral patterns were distinct from those of other Yemeni militant or tribal actors, we built a dataset of 576 newspaper articles drawn from the ICEWS corpus. Each article was coded for references to four actor categories: AQAP, Ansar al-Shariah, and the Houthi insurgency. We use newspaper articles rather than event dyads for evidence of behavioral convergence because newspaper reporting covers both specific activities as well as more diffuse framing. This duality provides the random forest with more opportunities to identify subtle differences and similarities. Each article was represented as a TF–IDF weighted bag-of-words vector comprising 2,222 terms common to all documents. During data preparation, we identified and removed a custom set of stopwords that served as class descriptors, such as sectarian identifiers, specific geographic identifiers, or significant personalities.  The full list of these stopwords is available in \nameref{S3_news}.  

 \underline{Random Forest And Diagnostic Interpretation}\\

Our primary classifier is a random forest, implemented in the randomForest package in R~\cite{breiman2018package}. Classification is complicated by class imbalance: the data contains relatively few articles on Ansar al-Shariah (36 of 576). To account for this, we stratified the model such that each tree was trained on a balanced subsample of 20 observations per class. To maximize interpretability, we intentionally sought models in a relatively simple parameter space.

To identify the final random forest model, we performed a 5-fold stratified cross-validation over 15 hyperparameter combinations (mtry $\in \{3, 5, 10, 15, 20\}$ and nodesize $\in \{1, 5, 1\}$), selecting optimal parameters by mean out-of-bag (OOB) error.  The stratified fivefold cross-validation was implemented using the caret package’s createFolds() function~\cite{kuhn2020package}.

Across all candidate models, the models successfully differentiated articles about the two Sunni insurgency organizations from those describing the Houthi insurgency. The final model, trained on all 576 articles and using the parameters that minimized out-of-bag error,
achieved high accuracy (Sensitivity: $\approx0.85$, Precision $\approx0.91$) for documents reporting on AQAP and Houthi activities but much weaker classification accuracy for reporting about the activities of Ansar al-Shariah (Sensitivity $\approx0.42$, Precision $\approx0.33$). Per-class performance metrics are reported in Table~\ref{class_metrics} with summary performance presented in Table~\ref{rfresults}.

\begin{table}[!ht] 
\centering 
\begin{tabular}{|l|l|l|l|} 
\hline 
\textbf{Class} & \textbf{N} & \textbf{Sensitivity} & \textbf{Precision}\\ \thickhline
\hline
Ansar al-Shariah & 36 & 0.42 & 0.33 \\ 
\hline
AQAP/Al-Qaeda & 327 & 0.86 & 0.93 \\
\hline Houthi/Ansarallah & 213 & 0.98 & 0.92 \\ 
\hline 
\end{tabular} 
\begin{flushleft} 
\caption{{\bf Classification Performance as Diagnostic Evidence for Theoretical Expectations}}
\label{class_metrics} 
Model performance metrics demonstrate diagnostic interpretation. Systematic inability to separate theoretically linked entities (AQAP and Ansar al-Shariah: 0.42 sensitivity) provides evidence of behavioral convergence, while successful differentiation of unrelated actors (Houthi: 0.98 sensitivity) validates the model's discriminative capacity. In this case, the model successfully distinguishes the unrelated civil war actors (AQAP and Houthi insurgencies) but has poor separation between the two linked organizations (AQAP and Ansar al-Shariah). We interpret this through the lens of our motivating theoretical expectations and as indicative of behavioral convergence in external perception of the two al-Qaeda linked militias.
\end{flushleft} 
\end{table}

\begin{table}[!ht]
\centering
\begin{tabular}{|l|l|l|l|}
\hline
\textbf{Metric} & \textbf{Value} & \textbf{SD} & \textbf{N} \\ \hline
\thickhline
\textbf{In-bag Accuracy}          & 0.88 & NA & 576 \\ \hline
\textbf{In-bag Balanced Accuracy} & 0.75 & NA & 576 \\ \hline
\textbf{OOB Error}                & 0.12 & NA & 576 \\ \hline
\textbf{CV Mean Accuracy}         & 0.88 & 0.03 & 576 \\ \hline
\textbf{CV Mean Balanced Accuracy}& 0.74 & 0.05 & 576 \\ \hline
\end{tabular}
\begin{flushleft}
\caption{{\bf Summary of Performance Metrics}}
\label{rfresults}
 Summary performance metrics for the final random forest model, showing across-class accuracy, error, and cross-validation. NA = not applicable for in-bag estimates. 
\end{flushleft}
\end{table}

\begin{figure}[!h]
\centering  
\includegraphics[width=\columnwidth]{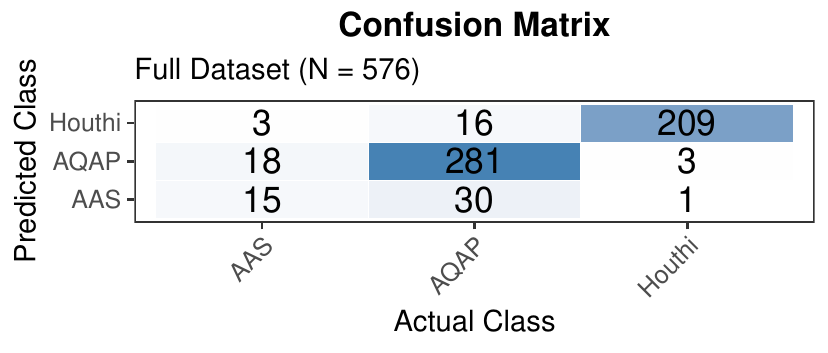}
\caption{{\bf Diagnostic use of classification confusion to detect behavioral convergence in opaque organizations.}
Full-model confusion matrix, showing strong ability to differentiate Houthi stories and convergence in reporting of al-Qaeda and Ansar al-Shariah activities. This supports the expectation that AQAP was unable to retain a fully separate positioning from their local spin-off.}
\label{fig:cm}
\end{figure}

Cross-validated results indicated high recall for Houthi-related articles ($0.98 \pm 0.02$) but much lower recall for Ansar al-Shariah ($0.36 \pm 0.14$), consistent with the expectation that media framing of AQAP and its local affiliate overlapped substantially. The model's strong ability to distinguish articles about the Houthi insurgency suggests that the model fails on the al-Qaeda groups because they are, indeed, difficult to distinguish from each other.  The confusion matrix in Fig~\ref{fig:cm} provides a clear diagnostic signal of convergence.

\begin{figure}[!h]
\centering  
\includegraphics[width=\columnwidth]{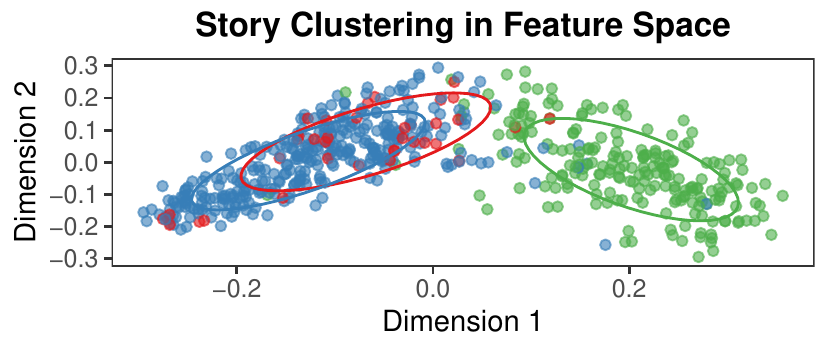}
\caption{{\bf MDS Projection.}
Projection of Yemen stories in feature space showing separation between Houthi (green) and Sunni-insurgent (red and blue) clusters and convergence between AQAP (blue)  and Ansar al-Shariah (red) texts.}
\label{fig:mds}
\end{figure}

A multidimensional scaling (MDS) projection of the random forest proximity matrix, pictured in Fig~\ref{fig:mds}, underscores the finding that reporting about the two jihadi groups with the same lineage (AQAP and Ansar al-Shariah) are virtually indistinguishable in feature space. Despite the official claims of an arms-length relationship and separate operation, in practice the two operated similarly. Conversely, even though they were also violent operators in the same territories and the same time period, articles about Houthi activities are clearly differentiable.  

Feature importance analysis via Mean Decrease in Gini highlighted terms relating to insurgency--- \textit{rebels, sanaa, saada, militants, and security}--- as the top five most important terms. This suggests that the al-Qaeda groups were jointly presented in one frame, namely as ``militants,'' with the implication of actors able to produce violence but not as an existential threat to the state while the Houthis were framed as ``rebels,'' a category of actor that captures their ability (and, in fact, success in) overthrowing the state.

\underline{Structural Topic Model and Temporal Patterns}\\

As a third triangulation step after showing convergence in external perception and behavior, we now ask whether AQAP’s own self-presentation exhibits the same localizing shift. We model expected changes in AQAP's self-presentation over time using the Structural Topic Model (STM)~\cite{roberts2014stm}. The STM is well-suited to analyzing longitudinal variation because it incorporates document-level metadata, such as publication date, into the estimation process.  Prior research has applied STM to corpora of short and moderate-length texts, including open-ended surveys~\cite{tingley2017rising}, social media messages~\cite{bail2016cultural}, and online forum posts~\cite{munksgaard2016mixing}. While general temporal trends cannot directly test the predictions of the bottom-up transformation mechanism, relative differences in topic prevalence can indicate its plausibility. In this context, increases in Yemeni topics and decreases in transnational or pan-jihadi topics would be consistent with an influential parochial base. Localizing pressures should manifest as a rising prevalence of themes tied to domestic fissures—such as the Houthi–Sunni civil war—and a decline in transnational themes, including references to other jihadi fronts and non-Yemeni targets. Such a shift would directly contravene al-Qaeda Central’s strategic advice to maintain focus on transnational conflict and avoid local entanglement.

To identify localization in self-presentation, we computationally analyzed a corpus of 809 documents issued by AQAP from June 2004 through September 2016. Official messages are an informative source for examining changing organizational priorities: within ideological and stylistic limits, these documents provide a venue through which the organization decides how to frame its public presentation. Furthermore, the technological environment makes propaganda documents an appealing source for analysis: since 2011, online platforms have been a major means of communication within Yemen~\cite[33]{carapico2014yemen}. This implies that media distributed online could have been consumed by both domestic and international audiences. The existence of such a dual audience likely discourages AQAP from strategically differentiating its online messaging from its local self-presentation. Discussion of corpus sourcing, translation, and potential biases is provided in Section~\nameref{S4_STM}.

To visualize these dynamics, we model the corpus with an 18-topic STM. Model specifications and robustness details are provided in the Supplementary Materials. The resulting topics were clustered into four thematic groupings: locally focused war reports, discussions of and threats concerning clandestine operations, promotion of transnational jihadi sentiments and goals, and jihadi-associated descriptors.\footnote{Three residual topics—Topics 1, 8, and 16—relate primarily to document construction and include terms associated with videography, standard sign-offs, and transcript production.} Across the corpus, the expected proportions devoted to local and transnational themes top out at less than half of hte corpus. This helps us in two ways, first, it indicates that both forms of rhetoric coexisted with each other and with other tertiary themes throughout the observation period. Secondly, and more importantly, because the local and transnational themes do not make up the entire corpus, we can infer drops in attention within the corpus as indications of reduced salience, rather than arithmetic crowding out. Summarized differently, if transnational themes decline over time, but the relative proportions of local and transnational themes never sum to 1, there remained rhetorical slack to increase the salience of transnational themes.  

\begin{figure}[ht]
\caption{\textbf{Topic clusters} Changes over time in attention dedicated to local and transnational themes. The clusters were generated from an 18-topic topic model. The topics evaluated and summarized by a subject matter expert who clustered topics that exhibit themes characteristic of a transnational jihadi focus, such as overthrowing regional governments and administering territory outside of Yemen, and those topics tied to domestic military and political events in Yemen.}
\label{fig:l-t-clusters}
\includegraphics[width=\linewidth]{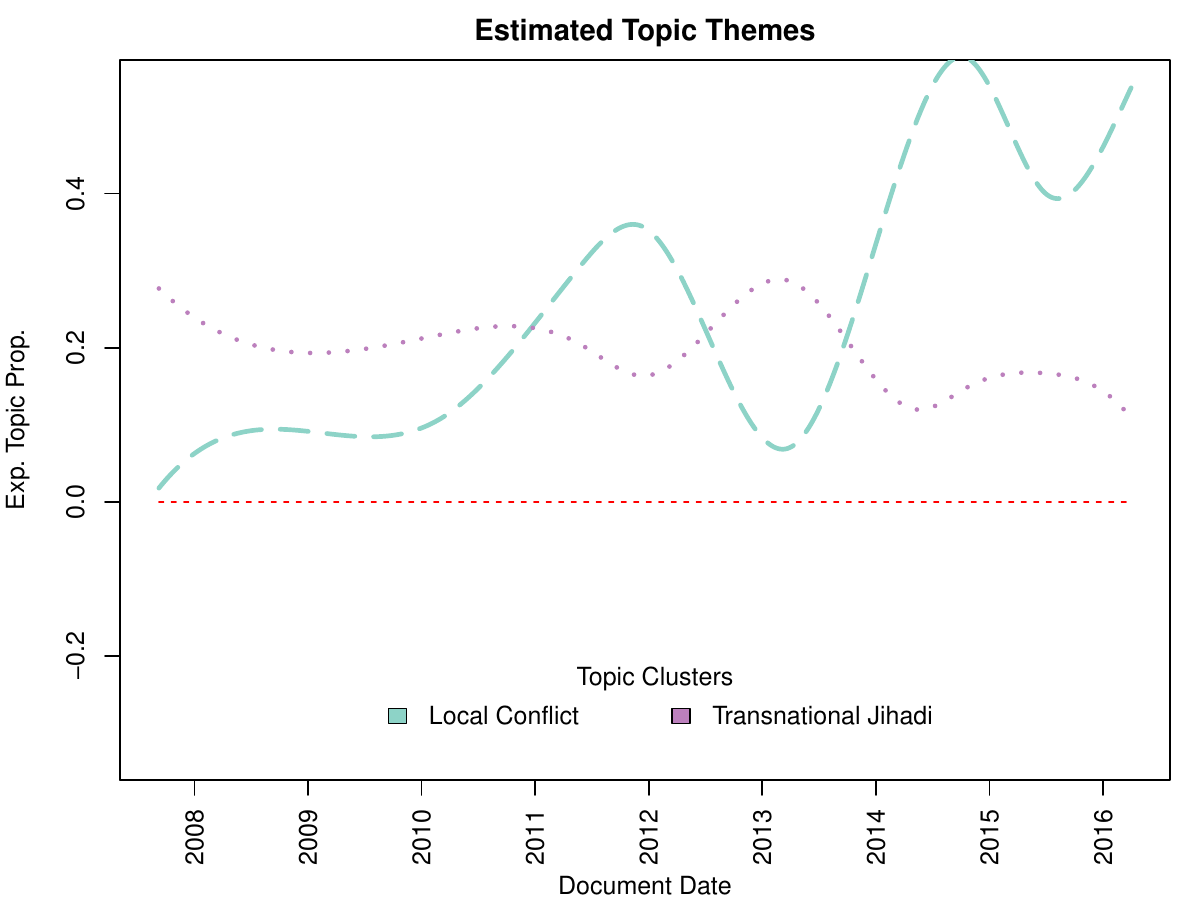}
\end{figure}

Figure~\ref{fig:l-t-clusters} presents the expected document-level proportions dedicated to local themes. These topics are characterized by terms referring to specific operations, geographic features, and political jurisdictions within Yemen. The prominence of these local conflict topics highlights a clear shift in attention toward domestic Yemeni issues. Conversely, the expected prevalence of transnationally focused topics—those addressing regional powers, such as the Saudi government and its security apparatus, and broader jihadi concerns—declines steadily after late 2012, as AQAP’s propaganda becomes increasingly focused on the Yemeni civil war. This pattern supports the accommodation scenario: local conflict topics rise while transnational themes decline after the 2011 expansion period, matching predictions that grassroots preferences would increasingly shape organizational messaging.

\section*{Discussion, Generalizability, and Extensions}

This manuscript has introduced a measurement framework using ML triangulation for opaque systems.  It enables bounded but defensible computational measurement in settings where the dynamics of interest are unobservable, data are fragmented, and the data-generating process for available signals is strategically concealed or missing from the record. Each of these elements poses a threat to standard inferential designs. The workflow presented here treats measurement as a structured process of cross-model evaluation across heterogeneous observables rather than as an attempt to recover hidden dynamics from a single data source or model. Our framework and workflow enable users to approach inhospitable data environments as constraints rather than insurmountable barriers. Moreover, by triangulating between theory and multiple interpretable ML designs, the framework shifts the target from recovering inaccessible ground truth to enabling systematic meso-level measurement in contexts when the available data is structurally adversarial.

This perspective suggests a shift from case-specific modeling toward reusable systematic measurement patterns for adversarial data. Although our demonstration focused on a single organization, the logic of systematic measurement developed here is portable to other domains characterized by opacity, including clandestine networks, complex organizations, and policy-relevant systems where direct observation is infeasible. Subsequent work can extend this framework by formalizing its epistemic assumptions and applying it across substantively distinct settings.

The measurement pattern developed above uses interpretable ML techniques to triangulate signals of internal processes from behavioral data that are publicly available. The approach is widely applicable to organizational settings characterized by:
\begin{enumerate}
\item{Limited internal visibility}
\item{Publicly available traces of organizational behavior, such as observable actions, news coverage, and statements}
\item{Theoretical expectations about structural pressures and how the organization should respond}
\end{enumerate}

This measurement strategy is particularly well-suited to domains where decision-makers must act despite persistent uncertainty rather than wait for identifiability.

\subsection*{Data Requirements and Potential Domains of Application}

Our approach is applicable across domains where difficult data regimes are a structural feature. The approach leverages a data-generating duality common to many important but difficult-to-study organizations.  Namely, organizations that strategically conceal their operations or operate in environments that do no produce consistent and granular data generate an adversarial data trail. Yet, success and impact drive outcomes and external interest which increases the quantity, quality, and variety of indirect data available to researchers. This combination creates the structural opportunity for data and model triangulation.

Although a flexible and effective strategy for measurement of varied and difficult-to-measure movements without demanding access to consistent fine-grained internal data, this approach requires access to:

\begin{enumerate}
\item{Theoretical priors: Specific expectations about what organizational change should look like are essential, as the theoretical framework enables validation of the computational measures. Moreover, the theoretical priors allow analytical leverage on when model failures represent a meaningful signal.}
\item{Multiple data pathways: At least two independent sources capturing different aspects of organizational behavior in order to triangulate from multiple angles. Examples of potential sources of data could include: internal communications and external perception or behavior and rhetorical output.}
\item{Temporal coverage: If measuring longitudinal processes, users must have time series coverage at the same scale as the target process. Our application focused on gradual change. Rapid changes should also be measurable, as long as they are matched by the  accessible data; however, it may require additional processing and analysis~\cite{foster2025subject}.}
\item{Tractable scale: Finally, the approach presented here scales effectively to corpora of thousands of documents while maintaining interpretability. Extension to massive datasets with millions of documents would require automation of the labeling and validation steps, though the core strategy remains applicable.}
\end{enumerate}

Types of organizations that fit this model could include:

\textit{Criminal organizations}: Cartels and trafficking networks leave traces in news coverage, law enforcement reports, and territory control patterns. The framework could detect shifts from profit-maximizing to protection-seeking behavior, or fragmentation following leadership decapitation.

\textit{Secretive corporations}: Privately held companies, firms in weakly regulated markets, or organizations accused of malfeasance often resist transparency. Public filings, news coverage, employee and job seeker social media platforms, and patent applications could reveal strategic pivots, cultural shifts, or responses to competitive pressure.

\textit{Political parties and movements}: Internal dynamics of political parties and social or political movements are frequently opaque, particularly in authoritarian contexts or during rapid mobilization. Rally speeches, social media, endorsement patterns, and campaign events provide observable traces of factional influence and strategic repositioning.

\textit{Online communities and platforms}: Extremist forums, cryptocurrency communities, or coordinated influence operations often operate semi-clandestinely. Platform data or saved participant records (where available) can be combined with external media coverage to track radicalization, coordination, or response to platform moderation.

\section*{Conclusion and Future Directions}

This manuscript presented a generalizable measurement framework for opaque systems with documented trajectories and big-picture trends, but where classical statistical and causal inference are unattainable because micro-level processes are structurally unobservable.  

We make three methodological contributions. First, we show how interpretable machine learning models can be deployed diagnostically: classification performance patterns become evidence about hidden processes. Second, we show how triangulating across heterogeneous data sources—such as external perception, self-presentation, and behavioral events—enables validation in the face of incomplete data access. Third, we provides a reusable workflow that practitioners can tailor for their research contexts, from criminal organizations to opaque corporations, without requiring specialized computational infrastructure or rare data access. Our framework supports theory-driven analysis in contexts where the big picture is known but available data do not support direct fine-grained inference, direct observation is impossible, and exact validation is unavailable.  

\underline{Future Directions}\\
 
 This manuscript represents an initial, applied, step toward a broader research agenda on inference in opaque and adversarial data regimes. Our work can be extended along several dimensions, covering variability of data, integration with forms of inference, and opportunities to scale the application. 

\textit{Additional measurement modalities}: Extensions of the framework could incorporate multi-modal data and ML models, widening the aperture. These extensions would broaden the types of opacity the framework can address. Similarly, network analysis could measure structural decentralization alongside behavioral evolution.

\textit{Transfer learning applications}: Training classifiers in one opacity context and testing generalization to related cases could improve efficiency and reveal common patterns across organizational types.

\textit{Integration with causal designs}: When natural experiments or instrumental variables become available, combining this measurement framework with causal identification strategies could move from characterization to causal estimation. The framework's measurements could serve as outcomes in quasi-experimental designs.

\textit{Automated quality metrics}: Another extension could introduce consistency scores across data sources or anomaly detection methods for unexpected divergence. This would enable application to new domains without extensive manual validation. This would improve scalability and reduce analyst workload.

\textit{Real-time monitoring applications}: Adapting the workflow for streaming data could enable early-warning systems detecting organizational shifts as they occur. This would be particularly useful for policy and intelligence applications that require a short lag time for detection.

\section*{Reproducibility and Adaptation}
All code is publicly available at \url{https://github.com/margaretfoster/growthtrap_rep}. Proprietary data, such as SITE Intelligence Group translations and ICEWS news reports, are available in processed format sufficient for replication.   The pipeline is implemented in R with both standard libraries (randomForest, stm) and custom extensions that are detailed in the GitHub repository. Results depending on statistical software with known breaking changes between versions, such as stm results, are included as both replication code and precomputed results. The documentation includes workflows, examples, and guidance for adapting to new organizational contexts, data sources, and research questions.

\section*{Declaration of generative AI and AI-assisted technologies in the writing process.}
 During the preparation of this work the author(s) used ChatGPT and Claude for developmental and copyediting assistance. After using this tool/service, the author(s) reviewed and edited the content as needed and take(s) full responsibility for the content of the published article.

 \section*{Declaration of Interests}
 The authors declare no competing interests.


\section*{Supporting Information}
\paragraph*{S1: Alternative Explanations and Sources of Bias}
\label{S1_bias}

The results presented are generally consistent with the theory's expectation that an influx of local fighters generated internal pressure on AQAP to focus on local issues. However, alternative explanations could account for the localizing pattern. The most notable of these explanations is that localization was a top-down strategic response rather than a bottom-up process of accommodation. The possibility of top-down strategic localization is significant because not only does it fundamentally challenge the accommodation mechanism, but the observable implications are the same as under the accommodation scenario. This section concludes by evaluating the potential for, and direction of, bias in the corpus and discussing additional possible cases beyond AQAP.

The most direct route for top-down strategic transformation would be for AQAP's leaders to have decided on a strategic change towards localization. Unfortunately, the opacity of organizational decision-making makes it particularly difficult to adjudicate between directed top-down change and gradual bottom-up accommodation, as doing so requires access to the inner workings of a secretive organization. However, AQAP's control and subsequent loss of territory in the Abyan Governorate in 2011 provides a rare window in which top leaders of the organization created documents illuminating their strategic thinking.

In 2013, the \textit{Associated Press} discovered a cache of documents left by al-Qaeda in the Islamic Maghreb (AQIM) in Timbuktu, Mali. Among the documents discovered by the \textit{Associated Press} were a pair of letters from \say{Abu Basir,} a nom de guerre of Nasir al-Wuhayshi, to his counterpart in North Africa, the leader of al-Qaeda in the Islamic Maghreb. The letters, which are dated from May and August 2012, feature al-Wuhayshi analyzing AQAP's administration of the population of Abyan and transmitting advice for AQIM's future governance. Although the comparison is not perfect--- the documents relate to governing rather than internal management--- they do provide a rare window into private strategic reflections.This paper focuses on AQAP's activities and messaging around 2011 and thus use the group's behavior during and after controlling territory in Abyan to develop observable implications from the counterfactual in which AQAP's localizing trend was a top-down strategic decision.

In recounting AQAP’s strategy for interacting with the population, al-Wuhayshi consistently presented a model of behavioral change that can be described as accommodation-to-radicalization. The first letter described a gradualist approach, writing \say{we have to first stop the great sins, and then move gradually to the lesser and lesser ones} and \say{our opinion in the beginning was to postpone the issue [of corporeal punishments during wartime]}~{\cite{al2012first}.  Rather than a logic of gaining strength by becoming more palatable to the communities and adopting their preferences, al-Wuhayshi proposed temporary leniency until jihadi administrators had time to indoctrinate the local community.

The impression that AQAP leaders intended accommodation to be temporary is underlined by the ratio of local to transnational topic proportions during and after 2011. While in control of Abyan in 2011 AQAP adopted a parochial stance, evident in the increase in local topics in Figure~\ref{fig:l-t-clusters}. However, shortly after relinquishing control of the governorate, AQAP returned to transnational jihadi themes. Notably, the largest bump in transnational content comes in 2013, the year after al-Wuhayshi's letter to AQIM that articulated their lessons from the failure to govern and retain Abyan. Retrenchment rather than a pivot towards local preferences is echoed in al-Wuhayshi's second letter, in which he observes that having to withdraw from Abyan provided an opportunity to consolidate via \say{a rare opportunity for guerrilla warfare and liquidations [assassinations]}~\cite{al2012second}. Al-Wuhayshi's letters contrast with a story of top-down strategic change, which could, for example, emphasize that integrating local priorities allowed for a more resilient presence.

This strategy is reflected in AQAP's later administration of the port city of Mukalla in 2015, in which AQAP attempted to introduce its governance through a conscious effort to co-opt local structures~\cite{alganad2020aqapadministration}.

Thus, AQAP's attempt to seize territory in Abyan Governorate in 2011 produced a specific moment of variation in localizing behavior and rhetoric. When in control of the southern governorate, AQAP attempted to administer the territory using their religious credentials as the basis for governance and political legitimacy~\cite{alganad2020aqapadministration}, but concluded that they would need a gradualist approach in the future. The failure of their initial governance experiment, the lessons that AQAP itself seems to have taken from the experience, and AQAP's subsequent rhetorical tilt away from local themes all point away from the alternative explanation of top-down adoption of local preferences.

A second alternative explanation could be that AQAP was simply following broader forces among the transnational jihadi community. One possible counter-narrative to the bottom-up transformation argument maintains that the change in al-Qaeda's leadership may have triggered a wider ideological shift that filtered to local branches, and that Awlaki's death amplified the effect in Yemen. This argument would undermine the argument that AQAP was divided between the preferences of their base and the explicit instructions of al-Qaeda Central.  Indeed, if al-Qaeda's central strategists changed their advice to local branches, changes in AQAP's self-presentation would not be informative about the grassroots transformation theory. Unfortunately, there are fewer captured strategic documents that directly attest to the strategic thinking of al-Qaeda's leaders in the second half of the data window. However, during this time, al-Qaeda Central leaders, particularly Emir Ayman al-Zawahiri, issued extensive public commentary and strategic advice for responding to regional upheavals such as the Arab Spring, the Syrian Civil War, and the al-Qaeda-Islamic State factional conflict. Thus, one can analyze communiques released by AQAP compared to those released by al-Qaeda's central propaganda mouthpiece, as-Sahab. This analysis indicates that as AQAP became more locally-focused, their messages increasingly diverge from propaganda released by al-Qaeda's central leadership. 

Using text to evaluate trends in self-presentation embeds the assumption that these documents present officially-sanctioned messaging. Although this assumption may be challenging in many contexts, al-Qaeda's online distribution networks have historically maintained very close oversight and control of their propaganda. Captured documents attest to AQAP's specific efforts to limit who can speak on behalf of the organization. Writing to a counterpart, Nasir al-Wuhayshi--- AQAP's leader until his death in June 2015--- indicated that AQAP limited the use of their brand, sharing \say{we restricted the statements and appearances of our brothers and emirs, allowing only those we deemed fit}~\cite{al2012first}.

\paragraph*{S2: Alternative Cases}
\label{S2_alternatives}

Readers may wonder whether the growth trap dynamic generalizes beyond the unique context of a local franchise of a transnational violent movement. The distinctive transnational and hierarchical structure of al-Qaeda and its local wings may lead readers to suspect that the expansion-transformation dynamic is more a story about the adaptation of a transnational ideology to a local context than it is about organizational pressure following from growth. Moreover, although AQAP presents a typical case in which to expect many of the theoretical drivers of the growth trap to be operative, lack of access to documentation about the inner workings of the group itself limits our ability to compare possible alternative explanations.

To address these worries, this section expands the aperture to briefly highlight three additional militant groups that have been subject to upward-driving pressure to change strategic focus. In each context, the literature describes militant groups that incorporated recruits with divergent preferences, exhausted their socialization capacity, and subsequently became reoriented to the preferences of the new base. The vignettes are selected to span geographic locations, time frames, and ideological backgrounds. Details on identification, selection, and other potential cases are featured in the Supplemental Materials.

The Sandinista National Liberation Front (FSLN) in Nicaragua illustrates how external community shocks can initiate bottom-up transformations. As \cite{mosinger2017dissident} details, in 1967 and 1972, ``grievance-triggering focus event[s]'' motivated new constituencies to regard the FSLN as a viable avenue through which to express anti-state grievances.  In 1967, the violent repression of a demonstration mobilized radical student organizations. Five years later, in 1972, government mismanagement of relief efforts after the Managua earthquake mobilized Christian activists. Recruits from the new constituencies then flocked to the FSLN and created new internal factions and external bases~\cite[210]{mosinger2017dissident}. Following both recruitment shocks, the FSLN was riven by internal power struggles as the new members sought to advance their preferences within the group. 

In 1968, the Palestine Liberation Organization claimed credit for fighting the Israeli Army to a stalemate in Karameh, Jordan. Reaping the rewards of a symbolic victory, the movement quickly gained thousands of new Palestinian and Arab recruits~\cite{sharif2009arafat}. However, this bounty rapidly turned toxic, as the new manpower quickly exceeded the PLO's absorption capacity, and the new fighters began abusing their host population in Jordan~\cite{szekely2017politics}. This abuse exacerbated tensions between the PLO and their Jordanian and Lebanese hosts, undermining the Executive Committee's strategic goal to remain on good terms with their sponsors~\cite{szekely2017politics}.

Seven years later, in 1975, a founder of the Eritrean Liberation Front (Jebha), Said Hussein, returned to the group after nine years in prison only to discover his organization transformed.  A nationalist group formerly dominated by conservative highland Muslims, the Jebha militia had been molded by an influx of Christians after Ethiopian crackdowns in 1974 and 1975. Indeed, after one crackdown, the number of prospective members so exceeded Jebha's absorption capacity that the group asked potential members to remain home until camp space opened~\cite[155]{woldemariam2016battlefield}. The new members, largely drawn from lowland Christian communities, quickly began pushing for Jebha to adopt a Marxist ideology anathema to the founders' socially conservative inclinations~\cite[111]{woldemariam2018insurgent}. 

The FSLN, PLO, and ELF represent a set of positive cases where scholars have independently observed a rapid growth to transformation dynamics. These provide initial support to the expectation that the growth trap mechanism is indeed one that affects militant groups in a variety of contexts and circumstances. Additional work can fully establish the scope of the phenomenon, as well as the counter-strategies that allow some militant groups to resist the dynamic outlined here.

\paragraph*{S3 Media Texts and Preprocessing}
\label{S3_news}

News stories originated in the ICEWS database and were selected by first querying the database for stories about events located in Yemen. This resulted in 47,385 stories ranging from January 15, 1991 through January 4, 2015.weselected only ``violent'' events, defined as events that fall into one of the following ICEWS event types: \textit{threaten with military force; use unconventional violence; violate ceasefire; use as human shield, threaten; occupy territory; physically assault; mobilize or increase armed forces; engage in violent protest for leadership change; engage in mass killing; conduct suicide; car, or other non-military bombing; carry out suicide bombing; attempt to assassinate, assassinate; abduct, hijack, or
take hostage; fight with small arms and light weapons; or fight with artillery and tanks}.

This produced 10,818 stories, of which we took a random sample of 1772 stories. ICEWS codes for event date, event type, source actor, and target actor. However, the source and target actor codes typically
characterize the actor by their role, such as \say{Armed Rebel} or \say{Militant} rather than by group affiliation. To generate data on how groups operate, we re-coded the reports to include a
variable for group or movement affiliation. We first sent the sample to Amazon's Mechanical Turk platform, asking the workers to categorize
the stories as relating to an action carried out by Ansar al-Shariah, AQAP, Houthi/Ansarallah, Yemeni Government, Tribal Uprising, Other,
Multiple Actors, or Unknown.  We kept the tags for the 283 stories that both coders agreed on, and hand-coded the remaining 1489 stories.  We then further subset the data to keep only the stories tagged as describing a violent event carried out by one of the three militias of interest. This produced the final 720-story corpus of news events. 
We evaluated a sample of 576 articles from the ICEWS media database that reported on events attributed to AQAP, Ansar al-Shariah, and the Houthi militias. We developed this sample by first creating a corpus of media reports of violent activity from Yemen for 2009-mid 2015 from the ICEWS database~\cite{boschee2015icews}. This generated 10,818 stories, covering November 1993 through January 2015. We randomly sampled 1,772 stories of violent activities and hand-coded the primary violent-producing actor featured in the selected articles.\footnote{Approximately 15\% of the data was coded by workers on Mechanical Turk, the rest was coded by the author.} 

From this set 566 articles described violence attributed to AQAP, Ansar al-Shariah, or the Houthi insurgency. We then randomly divided the stories into training (67\%) and test sets (33\%), yielding a training corpus with 432 articles and a test corpus with 144 articles. Both document sets spanned October 30, 2002 through January 3, 2015. Each tagged story was converted into a tokenized bag of words, normalized via term frequency-inverse document frequency (tf-idf). This produced a matrix with 2,222 tokens common to both sets. 

 We used the tm() package for R to tokenize the words in each story and remove numbers, standard English stopwords, whitespace, and stray HTML markup. We additionally removed a custom list of stopwords that strongly signal the group, such as variations on the group name and signifiers of sectarian identity. These custom stopwords are:\\
AQAP: qaeda, alqaida, alqaeda, qaida \\
Houthis: houthi, huthi, houthis, zaidi, alhouthi\\ 
Ansar al-Shariah: ansar, sharia, alsharia;\footnote{The robustness models also remove ``alshariah'' with little change in results.} \\
al-Qaeda: laden, osama; words often used to summarize location of action for one of the groups: peninsula, northern, southern, arabian, yemen[-]based\\
Sectarian identity: sunni, shia, shiite.\footnote{We did not remove areas of operation from the texts as the goal of the classifiers was to seek discussion of operational differences. Locations of operation are substantively meaningful.}

Word frequency was normalized via term frequency-inverse document frequency (tf-idf), producing a pair of tf-idf matrices, from which we took the intersection of features (i.e. words). This generated a set of 2,222 features for classification in the texts; which reduced the available terms significantly but was necessary to test models across the training, validation, and test sets.

We reattached metadata to each of the term document matrices. Metadata included group label, date, and whether the story was coded by Mechanical Turk workers. 
 
 We used this data to compare reporting of AQAP against it's own local spin-off organization, Ansar al-Shariah (Supporters of the Shariah).  Ansar al-Shariah was established in 2011 as an arms-length local wing that could focus on domestic grievances and administration rather than AQAP's transnational mission and which would be free of negative local sentiment associated with the al-Qaeda brand~\cite{ICG2017Yemen}. Although quickly identified as an alias for AQAP, having two different brands provides a reference point. Under the Ansar al-Shariah name, AQAP could strike a more parochial message, exploit local grievances, and avoid the encumbrances of the al-Qaeda brand~\cite{swift2012arc}.  In keeping with the expectation that local recruits would be primarily invested in the local conflict, many of these fighters \say{have deployed exclusively for an insurgency against the Yemeni government}~\cite[14]{hrw2013drone}.


\paragraph*{S4 Structural Topic Model}
\label{S4_STM}

Document identification and possible Sources of Translator Bias.  Readers may be concerned about the corpus origin, potential for bias, and risk of over-ascribing official approval to independently-created material. On the first concern: the corpus consists of English translations of Arabic releases and, occasionally, the original text of English-origin documents.\footnote{A very small portion of the documents, such as individual articles from \textit{Inspire Magazine}, were distributed in English.} These documents were collected and translated by the SITE Intelligence Group, a private research organization. These translations are advantageous for this project, as the company maintains near real-time coverage of prominent online distribution sites and has internal procedures to ensure consistent translation in style and tone.  The corpus is necessarily a sample, as to the author's knowledge, no comprehensive archive of all official AQAP propaganda exists in the public domain. However, any systematic selection effects should bias the results against finding increased local self-presentation, as the SITE Intelligence Group's document identification process is expected to prioritize documents accessible to an international audience. 

Readers may also worry that, even if online platforms are essential for communication in the country, Yemen's relatively low internet penetration rate suggests that AQAP's online propaganda is not intended for a domestic audience. If this is the case, it should likewise bias the results against findings in support of the bottom-up localizing hypothesis. A strategic actor could use in-person networks to signal which documents are intended for local versus international audiences. However, the difficult information and security environment in Yemen makes it risky to rely on a bifurcated media strategy. 

{\bf STM Preprocessing} We focus on media released online to jihadi media platforms and
outlets. From a corpus of professionally translated documents identified, translated, and edited by the SITE Intelligence Group, we selected all statements associated with AQAP from the start of the data collection (2004) until 2016. We kept those materials released in Arabic, omitting English language material such as Inspire Magazine, as that was produced for an external audience and therefore does not reveal local tension points. This produced a corpus of 809 candidate documents. Preprocessing removed words that occurred in fewer than two or more 70\% of the documents in the corpus.\footnote{In the AQAP corpus, there was
no change to the number of tokens in the corpus for an upper bound threshold between 70-95\%. We evaluated coherence and exclusivity at an upper threshold of 50\%, but did not find results that would suggest either a coherence or exclusivity benefit from the additional reduction in corpus size.} We dropped documents with too few words to model. The preprocessing resulted in a final corpus of 809 documents. 

The models used time as a covariate. The time variable is expressed in the data as a running counter of days from the oldest document in each corpus. Thus, the date of the oldest document is given as 1 and the day of each subsequent document is modeled as the number of days between the first date and the date of the individual document. 

These dates are linked to the translation date rather than release date as the former can be accurately pinpointed for each document in the corpus. For the vast majority of the corpus, the translation date closely coincides with the date that the document was released to online jihadi media outlets. The precision of release dates contained in the original Arabic text can vary according to type of document: communiques are typically dated to a specific day, while strategy documents or promotional magazine can be dated with a day, a month, or even a season. Thus, for consistency, the date covariate is linked to the translation date. Although not an exact map to the date of official release, SITE Intelligence Group had strong professional incentives to translate and release as close to real-time as possible.

{\bf Model Selection} Topic models rely on the user to prespecify a number of topics for the algorithm to search for. However, this parameter fundamentally influences the themes that will be identified in the documents.  For models one and two, we selected the number of topics by doing a
sweep of model specifications with 10 to 30 topics. We selected a topic number that performed best on both semantic coherence and exclusivity.\footnote{Ideally, the selected number of topics would have relatively high exclusivity and semantic coherence. We often
  faced a trade-off between the two. When determining the trade-off, we prioritized semantic coherence over exclusivity. In general, the exclusivity bands were narrow while coherence varied substantially.} After this process, a model with 18 topics appeared to present the greatest gains to semantic coherence without
trading off exclusivity. Moreover, the 18-topic model identified
topics that were particularly substantively coherent. 

As topic models are, by nature, non-deterministic, each implementation of a given model will produce slightly different results.  Thus, after selecting the number of topics for the STM to identify, we ran each model specification ten times to create a range of possible output models for analysis. We compared the average semantic coherence and exclusivity for each of the models. For each of the three models below, we found that the averages within each ten-model set were nearly identical. To avoid biasing my results by choosing output that best confirms my theoretical expectations, we selected from candidate models by maximizing average coherence and exclusivity metrics. As no model clearly dominated the coherence-exclusivity trade-off, we assigned a relatively stronger
weighting to semantic coherence when selecting a specific iteration to present. We selected a model before qualitatively evaluating any of the topics. This decision helps avoid bias in prioritizing coherence gains over exclusivity losses.\footnote{A plot of average semantic coherence and exclusivity scores is available upon request.} 
 
Finally, after selecting from among the candidate models, we evaluated the remaining models to ensure that the output was consistent across the set of ten results for each model. In particular, we verified general
agreement on the thematic content identified across the runs.

Readers may be concerned that the inverse relationships between topic proportion allocated to the \say{transnational} and \say{local conflicts} are simply mechanical. Although the total prevalence assigned to all topics in the model does sum to one, and thus increased attention to one topic necessarily means less attention to others, the \say{transnational} and \say{local conflicts} clusters never exceed an expected topic percentage of 75\% of any given document. 
When the entire corpus is taken together without any disaggregation by document release date, the two most common topics are related to the Houthi militias and terms that describe local targets and operations. Topics that associate words around ideological and tactical themes are each expected to feature in about 10\% of the total documents.
The Yemen topics identified in the main body of text group thematic clusters identified by the model relate to facets of AQAP's activities in Yemen, including discussion of Houthi militants, activities in Southern Yemen, castigation of the Yemeni government, and descriptions of local operations.  Documents representative of this topic are often battlefield communiques issued to claim local territorial control. 

\section*{Acknowledgments}

\end{document}